# Capturing global spatial context for accurate cell classification in skin cancer histology

Konstantinos Zormpas-Petridis[1*], Henrik Failmezger[2*], Ioannis Roxanis[3],
Matthew Blackledge[1], Yann Jamin[1], Yinyin Yuan[2]

[1] The Institute of Cancer Research, Division of Radiotherapy and Imaging, London, UK
[2] The Institute of Cancer Research, Division of Molecular Pathology, London, UK
[3] Royal Free London, NHS, UK
*E*-mail: konstantinos.zormpas-petridis@icr.ac.uk
yinyin.yuan@icr.ac.uk

**Abstract**

The spectacular response observed in clinical trials of immunotherapy in patients with previously uncurable Melanoma, a highly aggressive form of skin cancer, calls for a better understanding of the cancer-immune interface. Computational pathology provides a unique opportunity to spatially dissect such interface on digitised pathological slides. Accurate cellular classification is a key to ensure meaningful results, but is often challenging even with state-of-art machine learning and deep learning methods.

We propose a hierarchical framework, which mirrors the way pathologists perceive tumour architecture and define tumour heterogeneity to improve cell classification methods that rely solely on cell nuclei morphology. The SLIC superpixel algorithm was used to segment and classify tumour regions in low resolution H&E-stained histological images of melanoma skin cancer to provide a global context. Classification of superpixels into tumour, stroma, epidermis and lumen/white space, yielded a 97.7% training set accuracy and 95.7% testing set accuracy in 58 whole-tumour images of the TCGA melanoma dataset. The superpixel classification was projected down to high resolution images to enhance the performance of a single cell classifier, based on cell nuclear morphological features, and resulted in increasing its accuracy from 86.4% to 91.6%. Furthermore, a voting scheme was proposed to use global context as biological a priori knowledge, pushing the accuracy further to 92.8%.

This study demonstrates how using the global spatial context can accurately characterise the tumour microenvironment and allow us to extend significantly beyond single-cell morphological classification.

## 1 Introduction

Cell classification is an essential task in the histopathological characterisation of the tumour microenvironment. Differences in cell type abundance, regional distributions and spatial interactions can inform about the nature of disease, and provide robust markers of disease prognosis for risk-stratification [1]. In the new area of digital pathology, advanced image analysis can objectively, consistently and quantitatively characterise different components of the tumour and assist in tumour grading [2]. Accurate detection and classification algorithms are critical to assess the spatial distribution of all cell types.

Machine learning, and more recently, deep learning approaches have shown great promise in cell classification yielding high quality results [3,4,5]. However, even state-of-art algorithms tend to underperform in certain cases, as cell types often appear morphologically similar to each other or they overlap/touch. The existing pathological image analysis tools usually focus on individual cells' nuclei morphology with limited local context features, neglecting the tumour's global context.

Contextual information can be the key to further improve cell classification. Tumours are inherently heterogeneous, consisting of a mixture of tumour nests, lymphoid aggregates, stroma and other normal cell structures. Pathologists overcome the aforementioned issues by incorporating this context information and use tissue architecture, together with cell morphological features to accurately classify cells.

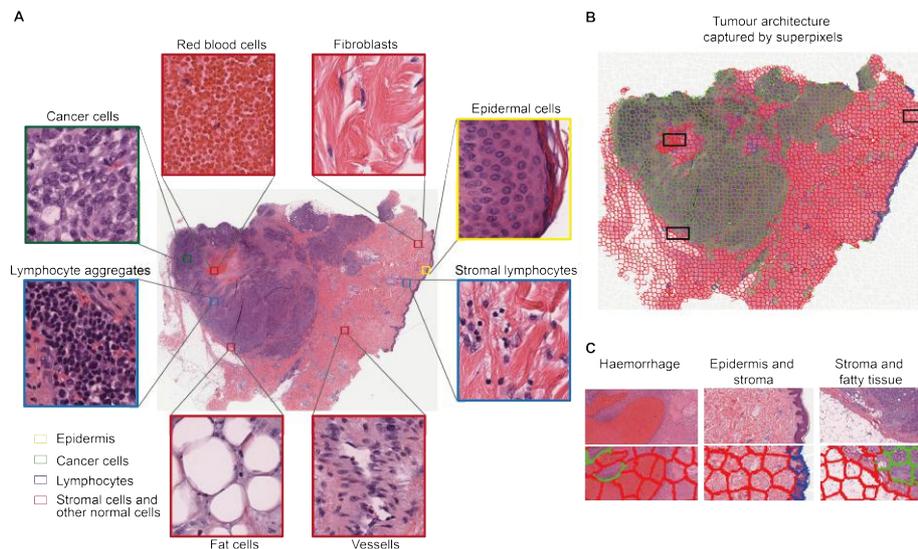

**Figure 1 The complex nature of melanoma architecture**. **A**. Heterogeneous tumour stroma makes accurate cell classification a difficult task. **B**. Superpixels captures tumour global architecture by delineating the boundaries among heterogeneous tumour components, including haemorrhage area, fatty tissues, stromal regions, epidermis and cancer nests. **C**. Current classification scheme assigned these components accurately to their respective superpixel classes. Top: example images; Bottom: segmentation and classification using SLIC superpixels.

Currently, there is a lack of methods to effectively define the boundaries of tumour components and propagate contextual features down, to aid cell classification. Proposed methods to separate tumour tissue into regions of interest (ROIs) include classification of image patches into diagnostically relevant ROIs using a visual bag-of-word model [6] and deep learning approaches [7]. Another popular method to capture similar local regions is the segmentation of image using superpixels [8]. *Bejnordi et al.* used superpixels to identify regions (stroma, background, epithelial nuclei) in whole-slide images (WSIs) at different magnifications [9]. Also, *Beck et al.* developed a framework that classifies superpixels as epithelium or stroma and used this framework in order to uncover stromal features that are associated with survival [10]. However, melanoma histology is highly heterogeneous, posing a number of challenges to machine learning such as class imbalance, intra-class diversity, and ambiguous tumour component boundaries (Figure 1A).

In this paper, we aim to overcome these challenges by effectively including a global tumour spatial context into single-cell classification. We propose a multi-resolution hierarchical framework, which captures the spatial global context at low magnification, by classifying superpixels into biologically meaningful regions (tumour area, normal stromal, normal epidermis and lumen/white space, Figure 1B-C) and combining them with cell nuclei morphological features at high resolution to improve single cell classification (Figure 2). We applied our algorithm on WSIs hematoxylin and eosin (H&E)-stained slides of melanoma skin cancer.

## 2   Methodology and Results

### 2.1   Data

58 full-face, H&E-stained section images from formalin-fixed, paraffin-embedded diagnostic blocks of melanoma skin cancer from the Cancer Genome Atlas were used. We scaled all digitized histology images to 20x magnification with a pixel resolution of 0.504μm using Bio-Formats (https://www.openmicroscopy.org/bio-formats/). To set the ground truth for regional classification, an expert pathologist provided annotations on the slides for 4 different regions: tumour area, normal stroma, normal epidermis and lumen/white space. We randomly selected 21 images for training and reserved the remaining 37 images as an independent test set.

For single cell classification, 7 WSIs (representative size: 30000x30000 pixels) were split into subimages (tiles) of 2000x2000 pixels each. 3 WSIs were used for training and 4 for testing. Based on pathologist's input, we used 3863 cell nuclei (1320 cancer cells, 1100 epidermal cells, 751 lymphocytes, 692 stromal cells) from 82 subimages for training and 2405 cell nuclei (876 cancer cells, 602 epidermal cells, 417 lymphocytes, 510 stromal cells) from 224 subimages as an independent test set (Figure 3A).

### 2.2   Superpixel classification

First, whole-slide full resolution images were downscaled to 1.25x magnification to retain overall tumour structures, while reducing the noise. Reinhard stain normalisation [11] was applied to account for stain variabilities that could affect the classification [12]. Subsequently, the images were segmented using the simple linear iterative clustering (SLIC) superpixels algorithm [8], which is designed to provide roughly uniform superpixels. Choosing the optimal number of superpixels is important to ensure that the superpixels capture homogeneous areas and adhere to image boundaries. With the pathologist's input, we visually identified a size of superpixels that met these criteria and chose the number of superpixels automatically based on each image's size (Eq. 1).

$$N_i = ceiling\left(\frac{S_i}{U}\right) \quad (1)$$

where $N_i$ is the number of superpixels in the $i$th image, $S_i$ is the size of image $i$ in pixels, and $U$ (here U = 1250) is a constant held across all images that defined a desired size of the superpixels. This means, on average, a superpixel occupies an area of approximately 35pixel-by-35pixel, equivalent to 280x280 micron. The SLIC superpixels algorithm was proven to be computationally efficient, requiring only 3s to segment a single downscaled image of 2500x2500 pixels using a 2.9GHz Intel core i7 processor.

We identified 15477 superpixels belonging in tumour areas, 6989 in stroma areas, 141 in epidermis and 691 in lumen/white space for training by determining whether their central points fell within the regions annotated by the pathologist.

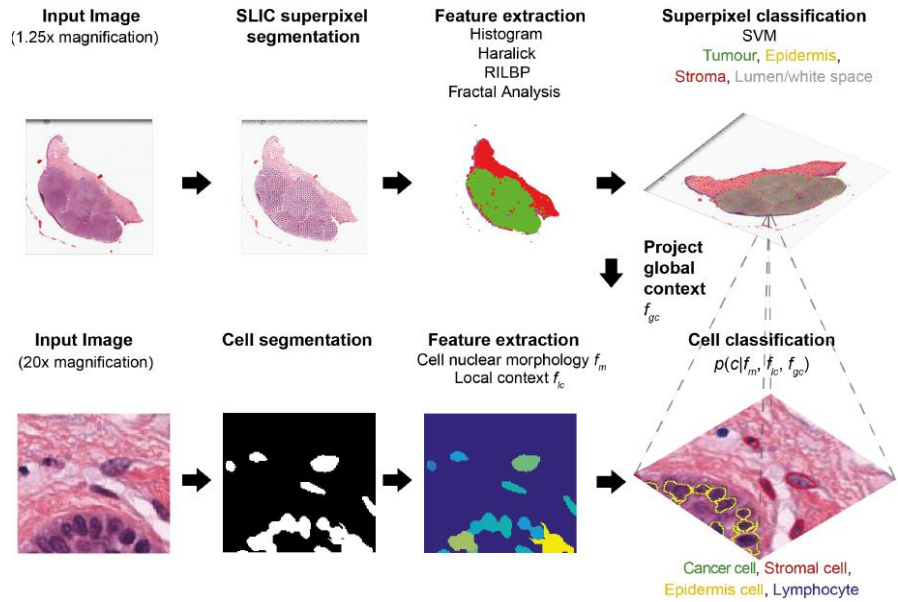

**Figure 2** Proposed hierarchical framework to project tumour global context onto single cell classification by integrating superpixel segmentation and classification.

Next, we extracted 4 types of features, totalling 85, from each superpixel, including 7 histogram features (mean value of hue, saturation and brightness, sum of intensities, contrast, standard deviation and entropy), and texture features (12 Haralick features [13], 59 rotation-invariant local binary patterns (RILBP), 7 segmentation-based fractal texture analysis (SFTA) features [14]). Features were standardized into z-scores. The mean values and SD of the features from the training set were used for the normalization of the test set. A support vector machine (SVM) with a radial basis function (RBF, γ=1/number_of_features) was trained with these features to classify superpixels into 4 different categories: cancer, stroma, epidermis and lumen/white space. To solve the class imbalance problem for training, we randomly selected a subset of 5000 cancer and stroma superpixels and increased the penalty in the cost function for the epidermis and lumen/white space classes by a factor of 10.

Performance of classification using individual and various combinations of feature sets was tested (Table 1). Using all 85 features, yielded the highest accuracy (97.7% in the training set using 10-fold cross validation and 95.7% in 2997 superpixels annotated in the 37 images of the independent test set).

**Table 1** Accuracy matrix of the superpixels' classification for single sets of features (left) and various combinations (right).

| Features | Accuracy (%) | Feature combinations | Accuracy |
|---|---|---|---|
| Hist. | 95.9 % | Hist. + Haralick | 97.3 % |
| Haralick | 91.4 % | Hist. + RILBP | 96.3 % |
| RILBP | 88.8 % | Hist. + SFTA | 96.9 % |
| SFTA | 85.2 % | Hist. + Haralick + RILBP | 97.1 % |
|  |  | Hist. + Haralick + SFTA | 97.1 % |
|  |  | Hist. + Haralick + RILBP + SFTA | **97.7** % |

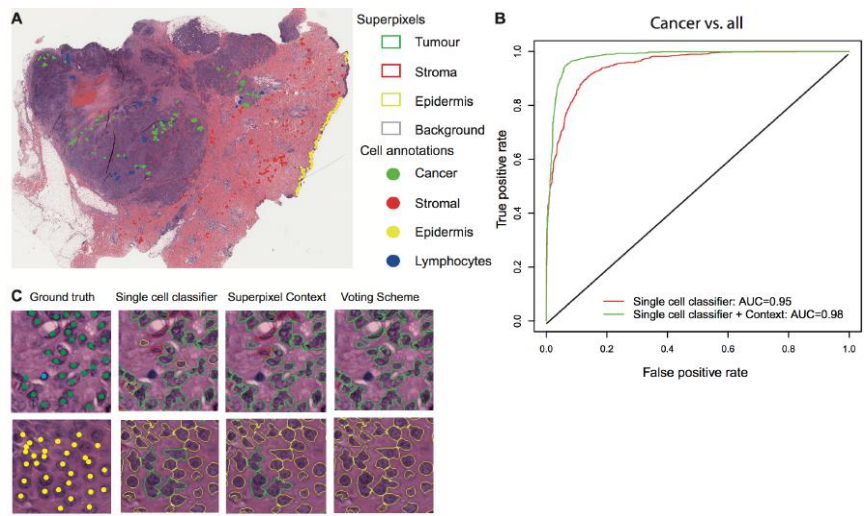

**Figure 3 Superpixels provide global context for single cell classification. A.** Representative superpixel classification of tumour regions overlaid with ground-truth cell annotations. **B**. ROC curves (cancer vs. all) illustrate the improvement in classification accuracy by adding superpixel context as additional features. **C**. Representative images comparing ground truth annotation and single cell classifiers with and without superpixels.

## 2.3   Cell Classification based on nuclear features

Image processing was carried out using the Bioconductor package EBImage [15]. Cell nuclei were extracted by Otsu thresholding; morphological opening was used to delete noisy structures in the image and clustered nuclei were separated by the Watershed algorithm. For every nucleus, 91 morphological features ($f_m$) were extracted [16].

Three local features $f_{lc}$ were added: the number of nuclei neighbours in a distance of 25 *μm*, the density at the particular cell position, and the size of the surrounding cytoplasm, calculated by thresholding the image's red channel after excluding the nuclei pixels.

For single cell classification, a Support Vector Machine (SVM) with a RBF (γ=1/number_of_features) kernel was trained with these features and achieved 86.4% accuracy (Table 3) in the test dataset. As expected, the classifier underperformed in distinguishing between epidermal and cancer cells (Table 2, Figure 3C), due to their similar morphology and related local context (both exist in crowded environments).

**Table 2** Confusion matrix of the Single Cell Classifiers. C: cancer, E: epidermis, L: lymphocyte, S: stromal. Red text highlight the confusion between cancer cells and epidermis cells.

|  |  | Morphology | | | | Morphology + global context | | | | Morphology + Voting scheme | | | |
|---|---|---|---|---|---|---|---|---|---|---|---|---|---|
|  |  | C | E | L | S | C | E | L | S | C | E | L | S |
| Classes | cancer | **715** | 127 | 20 | 14 | **786** | 27 | 23 | 40 | **844** | 0 | 30 | 2 |
|  | epidermis | 96 | **496** | 5 | 5 | 44 | **550** | 2 | 6 | 48 | **515** | 17 | 22 |
|  | lymphocyte | 7 | 12 | **397** | 1 | 12 | 5 | **399** | 1 | 14 | 1 | **401** | 1 |
|  | stromal | 18 | 4 | 17 | **471** | 11 | 12 | 18 | **469** | 10 | 4 | 23 | **473** |

### 2.4 Cell classification with Context

Two different schemes were used in order to integrate regional classification with cell classification. First, the type of area a single cell belonged to, provided by the superpixel classification, was added to the morphological feature set as the global context feature ($f_{gc}$). This reduced the misclassification between epidermis and cancer cells in large degree (Table 2) and led to a much higher accuracy (91.6%, Table 3, Figure 3BC) compared to the cell-morphology based classifier.

Secondly, global context given by superpixels served as biological *apriori* knowledge to correct single cell classifications. E.g. stromal cells seldom exist in non-stromal regions, while, cancer cells should only exist in tumour regions and epidermal cells should be found only in epidermal regions. Lymphocytes, however, can infiltrate into both tumour regions and stroma, but are rarely found in epidermis [17]. The regional context of a cell is thus of great importance for its annotation. We subsequently implemented an iterative voting scheme for cells in stromal and tumour regions:

$$c_i = \begin{cases} epidermis & if & s = epidermis \\ t & if\,else & t = s \vee t = lymphocyte \\ k & else & k = s \vee k = lymphocyte \end{cases} \quad (2)$$

Where $c_i$ is the cell at position $i$, $t \in \{cancer, stromal, epidermis, lymphocyte\}$ is the most probable annotation of the cell in the SVM, $s \in \{cancer, stromal, epidermis\}$ is the annotation of the cell's superpixel, $k \in \{cancer, stromal, epidermis, lymphocyte\}$ is the annotation with the next highest unchecked probability in the SVM. This voting scheme (Eq. 2) was applied for all cell nuclei and resulted in 92.8% accuracy in the test dataset (Figure 3C, Table 3).

**Table 3** Accuracy of the classifiers.

| Method | Accuracy | Precision | Recall |
|---|---|---|---|
| Single Cell | 86.4 % | 87.4 % | 87.9 % |
| Single Cell + Context | 91.6 % | 91.5 % | 92.2 % |
| Voting Scheme | 92.8 % | 92.8 % | 92.7 % |

## 3  Discussion

Accurate characterisation of the tumour microenvironment is crucial for understanding cancer as a highly complex, non-autonomous disease. In this study, we built a hierarchical framework to mirror the way pathologists perceive tumour architecture. Despite recent advances in image analysis, there are limitations associated with cell classification strictly based on their nuclear morphology or local contextual features, due to phenotypical similarities; this can be overcome by incorporating the global spatial context. Multi-resolution or superpixel-based methods have been successfully proposed for identifying ROIs [9] and unsupervised segmentation [18,19]. Here, our methodology enables the classification of cancer and diverse microenvironmental cells with high accuracy through the implementation of contextual features and a voting scheme.

We demonstrated that the addition of global context feature from superpixels improved cell classification based only on nuclei morphology from 86.4% to 92.8%. After the validation of our method on a larger dataset, we intend to combine it with more sophisticated cell segmentation and classification deep learning algorithms [5] and pave the way towards automatic scoring to assist in the stratification of melanoma patients. Such system can provide a better understanding of the cancer-immune cell interface, cell-stroma interactions and predictive biomarkers of response to novel therapies, including immunotherapy, which has radically changed melanoma patient survival. Also, the proposed framework can be easily adapted and used to study other cancer types.

## 4  Conclusion

We have demonstrated that our multi-resolution approach, incorporating spatial tissue context improves the accuracy of automated cell classification from digital histopathology compared to our conventional single cell classification methodology based solely on nuclear morphology in clinical melanoma samples.